\documentclass[conference]{IEEEtran}
\usepackage{cite}
\usepackage{amsmath,amssymb,amsfonts}
\usepackage{algorithmic}
\usepackage{algorithm}
\usepackage[colorlinks=true, linkcolor=blue, citecolor=red, urlcolor=cyan]{hyperref}
\usepackage{graphicx}
\usepackage{textcomp}
\usepackage{xcolor}
\usepackage{booktabs}
\usepackage{multirow}
\usepackage{url}
\usepackage{float}
\usepackage{stfloats}
\usepackage{afterpage}

\begin{document}

\title{Personalized QoE Prediction: A Demographic-Augmented Machine Learning Framework for 5G Video Streaming Networks}

\author{
\IEEEauthorblockN{Maryam Khalid}
\IEEEauthorblockA{
\textit{Department of Data Science and AI} \\
\textit{National University of Computer and Emerging Sciences} \\
Islamabad, Pakistan \\
i221917@nu.edu.pk
}
\and
\IEEEauthorblockN{Hijab Beg}
\IEEEauthorblockA{
\textit{Department of Data Science and AI} \\
\textit{National University of Computer and Emerging Sciences} \\
Islamabad, Pakistan \\
i222071@nu.edu.pk
}
\and
\IEEEauthorblockN{Zunaira Ahmed}
\IEEEauthorblockA{
\textit{Department of Data Science and AI} \\
\textit{National University of Computer and Emerging Sciences} \\
Islamabad, Pakistan \\
i222075@nu.edu.pk
}
\and
\IEEEauthorblockN{Instructor: Mohsin Khan}
\IEEEauthorblockA{
\textit{Faculty of Computing} \\
\textit{National University of Computer and Emerging Sciences} \\
Islamabad, Pakistan \\
Mohsin.khan@isb.nu.edu.pk
}
}

\maketitle

\begin{abstract}
Quality of Experience (QoE) prediction has become a critical component in modern multimedia and networked systems, especially as adaptive video streaming continues to dominate global internet traffic. Accurate and scalable QoE estimation enables intelligent resource management in 5G/6G networks and enhances user‐centric service delivery.

Previous studies, including the widely cited supervised‐learning tutorial for HAS-based QoE prediction, primarily focused on traditional machine learning models and relied on limited datasets with uniform user assumptions. These approaches, while effective, fail to capture demographic variability and deeper feature dependencies.

In this work, we introduce a demographic-aware data augmentation strategy combined with advanced deep learning techniques to improve QoE prediction robustness. We construct synthetic but behaviorally realistic demographic profiles and expand the dataset sixfold, enabling richer modeling of user perception diversity. On top of the engineered dataset, we evaluate a comprehensive set of classical ML models and state-of-the-art deep neural architectures. Among these, TabNet—an attentive, feature-selection-driven deep learning model—achieves the strongest performance, demonstrating superior interpretability and generalization on augmented QoE features.

Experimental results show significant performance gains across RMSE, MAE, and $R^2$ metrics compared to baseline models from prior work. Our findings confirm that demographic-aware augmentation and attention-based deep learning substantially enhance QoE prediction accuracy and robustness, offering a more realistic and scalable direction for future QoE-aware network intelligence.
\end{abstract}

\begin{IEEEkeywords}
Quality of Experience (QoE), Machine Learning, Data Augmentation, Video Streaming, User Demographics, Supervised Learning.
\end{IEEEkeywords}

\section{Introduction}
\label{sec:introduction}

The rapid growth of multimedia services and the widespread adoption of HTTP Adaptive Streaming (HAS) have made Quality of Experience (QoE) prediction an essential component of modern communication networks. As video traffic continues to dominate global Internet usage, achieving accurate, scalable, and user-centric QoE prediction has become vital for efficient resource management, service optimization, and the deployment of QoE-aware mechanisms in emerging 5G and 6G systems.

Previous research has extensively examined QoE prediction through supervised machine learning models. Among these, the work of Ahmad \textit{et al.} provides a comprehensive tutorial and comparative study for HAS-based QoE prediction, including a complete pipeline from data collection to model evaluation. While such approaches demonstrate strong performance using methods like Random Forest and Gradient Boosting, they are built on limited datasets and assume that all users perceive streaming impairments similarly. This uniform modeling of users restricts the applicability of QoE prediction models to diverse real-world scenarios.

Current QoE datasets are typically small and lack representation of demographic diversity, which results in models that generalize poorly when deployed in heterogeneous environments. Moreover, existing approaches focus primarily on classical machine learning methods, leaving a gap in leveraging more advanced deep learning architectures capable of capturing complex nonlinear relationships in QoE-relevant features. These limitations highlight the need for a more comprehensive and realistic modeling strategy that accounts for the variations in human perception across different demographic and behavioral patterns.

To overcome these challenges, this work introduces a demographic-aware QoE prediction framework. The proposed methodology begins by constructing behaviorally realistic demographic profiles that model different levels of user sensitivity to key QoE factors such as stalling events, bitrate variations, and visual degradation. These profiles are then used to augment the original QoE dataset sixfold through a targeted transformation of the subjective MOS scores, generating a richer and more diverse dataset that reflects real differences in user perception. Building upon this enhanced dataset, we evaluate a wide range of classical machine learning models alongside modern deep learning architectures. In particular, attention-based networks and the tabular deep learning model TabNet are examined for their ability to learn complex feature dependencies and perform automatic feature selection. Models are compared using standard performance metrics, including RMSE, MAE, and $R^2$, to determine their effectiveness in predicting QoE under the new demographic-aware paradigm.

\subsection{Key Contributions}

The main contributions of this research are as follows:

\begin{itemize}
\item \textbf{Demographic-Aware Dataset Expansion:} Development of six behaviorally realistic demographic profiles that capture variability in user sensitivity, expanding the original QoE dataset from 450 to 2700 samples.

\item \textbf{Perception-Based MOS Transformation:} Introduction of a demographic-driven MOS adjustment function that simulates realistic variations in QoE perception across different user groups.

\item \textbf{Unified Evaluation of Classical and Deep Learning Models:} Comprehensive benchmarking of traditional ML algorithms alongside advanced deep learning models, including attention-based architectures and TabNet.

\item \textbf{Superior Performance of TabNet:} Demonstration that TabNet achieves the best predictive accuracy among all tested models due to its inherent feature selection and attention mechanisms.

\item \textbf{Improved Generalization and Practicality:} Evidence that the proposed demographic-aware augmentation significantly enhances model robustness, making QoE prediction more reflective of real-world user diversity.
\end{itemize}

\section{Related Work}
\label{sec:related_work}

Research on Quality of Experience prediction has grown significantly in recent years. Early studies focused on linking network performance metrics to user satisfaction. For example, the work by Duanmu et al. \cite{duanmu2018} created a database for adaptive video streaming by collecting both objective network parameters and subjective user scores. Their approach showed that simple regression models could map technical features to Mean Opinion Scores with reasonable accuracy. However, their study treated all users as a single homogeneous group, averaging individual differences which limits personalization.

Another important contribution came from Ahmad et al. \cite{ahmad2021}, who provided a comprehensive tutorial on supervised learning for QoE prediction. They compared various machine learning algorithms including Random Forest, Support Vector Machines, and neural networks. Their results indicated that ensemble methods like Random Forest performed best on standard datasets. While their work established a solid pipeline for QoE prediction, it did not address how different user types might perceive the same network conditions differently. This limitation is particularly relevant for 5G networks where diverse applications require personalized quality management.

Standards organizations have also developed frameworks for quality assessment. The ITU-T P.1203 recommendation \cite{itu_p1203} provides parametric models for estimating video quality based on bitstream information. These models are useful for network operators because they do not require access to the original video. However, these standardized models are designed for general use and lack the flexibility to adapt to specific user preferences or demographic variations. They apply the same quality thresholds to all users regardless of their individual sensitivities.

Recent surveys have highlighted the need for more personalized approaches. Barman and Martini \cite{barman2019} conducted a comprehensive review of QoE modeling for HTTP adaptive streaming. They identified that most existing models fail to account for user diversity, treating all viewers as having identical expectations. Their survey pointed out that future research should incorporate demographic factors and usage contexts to improve prediction accuracy. This observation directly motivated our work on demographic aware augmentation.

Some researchers have attempted to incorporate user context into QoE models. Vega et al. \cite{vega2018} reviewed predictive QoE management systems and noted that context awareness was an emerging trend. They discussed how factors like device type, viewing environment, and user activity could influence perceived quality. However, most implementations they reviewed relied on simple rule based adjustments rather than learning personalized patterns from data. This gap suggests a need for more sophisticated approaches that can learn complex user behavior patterns.

Deep learning approaches have been explored in related fields. For example, Osa et al. \cite{osa2024} applied neural networks to intrusion detection in networks, demonstrating their capability to learn complex patterns. However, their work required large datasets for effective training, which are not typically available for QoE research due to the cost and difficulty of collecting subjective user ratings. This data scarcity problem has limited the application of deep learning in QoE prediction, creating a need for techniques that can work with smaller datasets.

Network slicing and resource allocation research has also considered QoE. Piamrat et al. \cite{piamrat2021} discussed QoE aware resource allocation for video streaming in 5G networks. Their work emphasized the importance of dynamic resource distribution based on user needs. However, their approach assumed uniform QoE requirements across users, which does not reflect real world diversity. Our work extends this concept by enabling personalized QoE predictions that can inform more nuanced resource allocation decisions.

In summary, while existing research has made significant progress in QoE prediction, several limitations remain. Most studies treat users as homogeneous groups, use small datasets that cannot support complex models, or rely on rigid standardized models that cannot adapt to individual differences. Our work addresses these limitations by introducing demographic aware data augmentation that expands small datasets while preserving behavioral diversity, enabling more personalized QoE predictions suitable for 5G network slicing scenarios.

\section{Proposed Architecture and Case Study}
\label{sec:architecture}

We wanted to make sure our research had some practical use, so we designed everything around this 5G Network Slicing scenario that's actually becoming relevant now.

\subsection{Case Study Scenario}

Picture a typical urban area with a 5G small cell serving all kinds of users at once. For our study, we focused on three main groups that we thought represented common situations:

First you've got the gamers - these people playing competitive games where every millisecond of delay matters. Then there's commuters watching videos on their phones while traveling - they're usually okay with lower quality as long as it doesn't keep freezing. And finally you've got the more casual users, like elderly people watching TV shows on smart displays at home - they mostly just want something that works consistently.

The idea is to have this QoE prediction engine running at the network edge that can monitor what's happening in real-time and predict MOS scores for each user type. So if it notices that a gamer's experience is about to go downhill, it can quickly give them more low-latency resources, maybe even taking some bandwidth from users who wouldn't notice the difference as much.

\section{Architecture}
\label{sec:architecture_detail}

\begin{figure*}[t!]
\centering
\includegraphics[width=\textwidth]{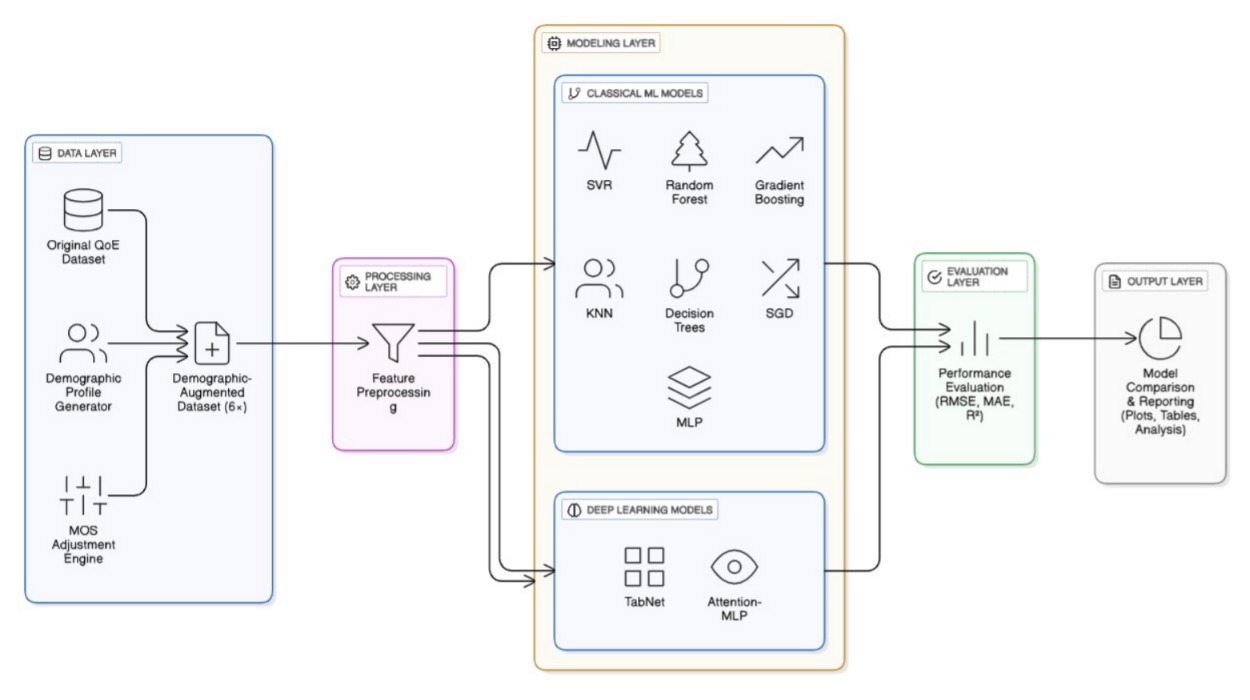}
\caption{Architecture Diagram}
\label{fig:architecture}
\end{figure*}

The system architecture for the proposed demographic-aware QoE prediction framework is organized into five functional layers, each representing a distinct stage of data processing, modeling, and evaluation. The architecture is illustrated in Figure~\ref{fig:architecture}, where components are grouped according to their role in the overall workflow.

\textbf{1. Data Layer:} The architecture begins with the original QoE dataset, which contains video session features and corresponding subjective quality scores. To address the limitation of user uniformity, a demographic profile generator is introduced to create multiple user profiles with varying sensitivities to quality impairments. These profiles serve as input to the MOS adjustment engine, which modifies the original subjective scores based on realistic behavioral patterns. This process expands the dataset sixfold, producing a demographic-augmented dataset that captures diverse perception characteristics.

\textbf{2. Processing Layer:} The augmented dataset is passed through a feature preprocessing module responsible for normalization, cleaning, and preparing the data for model training. This stage ensures uniform scale, reduces noise, and handles feature transformations required for both classical and deep learning approaches.

\textbf{3. Modeling Layer:} The preprocessed data is then fed into two parallel modeling pipelines. The first pipeline contains classical machine learning algorithms such as Random Forest, Gradient Boosting, Support Vector Regression, KNN, Decision Trees, SGD, and MLP. The second pipeline employs advanced deep learning architectures, specifically an attention-based MLP and the TabNet model, which leverage feature-selection and adaptive attention mechanisms to learn complex QoE patterns.

\textbf{4. Evaluation Layer:} Both sets of models are evaluated using standard regression metrics, including RMSE, MAE, and the coefficient of determination ($R^2$). This layer provides a consistent and objective comparison of model performance across the expanded demographic-aware dataset.

\textbf{5. Output Layer:} The final layer performs comprehensive model comparison and reporting. It aggregates evaluation results into plots, tables, and analytical summaries. This stage provides insights into model behavior, highlights performance differences between classical and deep learning approaches, and demonstrates the impact of demographic-aware augmentation on QoE prediction.

\section{Methodology}

This section explains the complete methodology used to develop a demographic-aware QoE prediction framework. The process contains five main stages: (1) dataset preparation, (2) demographic profile design, (3) MOS augmentation, (4) preprocessing, and (5) model training and evaluation. Each step is directly based on the implementation described in the accompanying source code.

\subsection{Base Dataset}

The starting point of the methodology is the original HAS-based QoE dataset containing 450 video streaming sessions. Each sample includes objective streaming metrics such as VMAF, SSIM, bitrate statistics, rebuffering duration, QP values, and device or content metadata. The original dataset also provides a Mean Opinion Score (MOS), which serves as the target variable.

Let each streaming session be represented as:
\[
x_i = (f_{i1}, f_{i2}, \ldots, f_{id}), \quad y_i = \text{MOS}_i
\]
where $f_{ij}$ denotes the $j$-th feature and $y_i$ is the human-reported MOS.

\subsection{Demographic Profile Design}

To overcome the lack of demographic diversity in the original dataset, six synthetic user groups are defined. Each demographic group is associated with sensitivity weights for different QoE factors:
\[
\text{Profile}_k = \{ w^{(k)}_{\text{rebuff}}, \ w^{(k)}_{\text{quality}}, \ w^{(k)}_{\text{bitrate}}, \ w^{(k)}_{\text{consistency}} \}
\]

Each profile also includes a specific MOS adjustment function that modifies the original MOS score.

The six profiles implemented are: \textit{casual viewer, quality enthusiast, mobile user, gamer/sports viewer, elderly user, and professional-critical user.}

\subsection{Impact Factor Computation}

For each streaming session, the system computes QoE impact factors from raw features. These factors express rebuffering impact, quality boost, quality variance, smoothness, simplicity, and compression level.

Key factors include:
\[
\text{RebuffImpact} = \min\left(\frac{\text{stall\_duration}}{2.0}, \ 1.0\right)
\]
\[
\text{QualityBoost} = \frac{1}{2}\left( \frac{\text{VMAF}}{100} + \text{SSIM} \right)
\]
\[
\text{QualityVariance} = \frac{1}{2}\left( \frac{\sigma_{\text{VMAF}}}{\mu_{\text{VMAF}}} + \frac{\sigma_{\text{bitrate}}}{\mu_{\text{bitrate}}} \right)
\]
\[
\text{Smoothness} = 1 - \min(\text{QualityVariance}, 1)
\]

These factors modify user perception differently for each demographic group.

\subsection{Demographic-Based MOS Augmentation}

For every original sample, the system generates six augmented versions (one per demographic). The MOS adjustment is computed using the demographic-specific function:
\[
\widehat{y}_{i,k} = g_k\left(y_i, \ \text{ImpactFactors}(x_i)\right)
\]
where $g_k(\cdot)$ is the adjustment rule defined for demographic $k$.

To simulate natural variation among individuals, Gaussian noise is added:
\[
\widetilde{y}_{i,k} = \widehat{y}_{i,k} + \epsilon, \quad \epsilon \sim \mathcal{N}(0, \sigma^2), \ \sigma = 2.0
\]

Finally, scores are clipped into the valid MOS range:
\[
\widetilde{y}_{i,k} = \min(\max(\widetilde{y}_{i,k}, 0), 100)
\]

This augmentation increases the dataset size from 450 to 2700 samples.

\subsection{Feature Preprocessing}

Before model training, several preprocessing steps are applied:

\begin{itemize}
\item Label-encoding of categorical variables (content, device, encoding profile)
\item Removal of non-predictive fields such as log paths
\item Standardization of all numerical features using:
\[
x' = \frac{x - \mu}{\sigma}
\]
\end{itemize}

This ensures compatibility with both classical ML models and deep learning networks.

\section{Model Training}

This section explains the training procedure for all classical machine learning models used in the study. Each model was trained on both the original dataset and the demographic-augmented dataset. The goal of the training process was to learn a mapping from streaming metrics to predicted MOS values. Let the training dataset be represented as:
\[
\mathcal{D} = \{ (x_i, y_i) \}_{i=1}^{N},
\]
where $x_i$ is the feature vector for session $i$ and $y_i$ is the corresponding MOS score. All models attempt to learn a function $f(\cdot)$ such that:
\[
\hat{y}_i = f(x_i).
\]

Before training, all numerical features were standardized, and categorical features were label-encoded. The training set was used to fit the internal parameters of each model, while the test set was held out for later evaluation.

\subsection{Linear Regression}

Linear Regression assumes a linear relationship between input features and MOS. The model estimates a weight vector $\beta$ such that:
\[
\hat{y} = \beta_0 + \beta_1 x_1 + \beta_2 x_2 + \cdots + \beta_d x_d.
\]

The weights are learned by minimizing the squared error between predictions and ground truth. This model serves as a simple baseline.

\subsection{Decision Tree Regressor}

A Decision Tree Regressor partitions the feature space into regions and fits a constant MOS value in each region. It splits the data using thresholds that reduce impurity in the target variable. For each split, the model chooses the feature $j$ and threshold $t$ that minimize:
\[
\text{Loss} = \sum_{x_i \in R_1(j,t)} (y_i - \bar{y}_{R_1})^2 + \sum_{x_i \in R_2(j,t)} (y_i - \bar{y}_{R_2})^2.
\]

This allows the model to capture nonlinear relationships between features and MOS.

\subsection{Random Forest Regressor}

Random Forest is an ensemble model that trains multiple decision trees on random subsets of the data and features. Each tree produces a prediction, and the forest outputs the average:
\[
\hat{y} = \frac{1}{T} \sum_{t=1}^{T} f_t(x),
\]
where $T$ is the number of trees. This reduces variance and improves generalization. Hyperparameters include the number of trees and tree depth.

\subsection{Gradient Boosting Regressor}

Gradient Boosting builds trees sequentially, where each new tree tries to correct the errors of the previous ones. For iteration $m$, the model fits a new tree $h_m(x)$ to the residuals:
\[
r_i^{(m)} = y_i - \hat{y}_i^{(m-1)}.
\]

The updated model is:
\[
\hat{y}_i^{(m)} = \hat{y}_i^{(m-1)} + \eta \, h_m(x_i),
\]
where $\eta$ is the learning rate. This allows the model to learn complex patterns gradually.

\subsection{Support Vector Regression (SVR)}

SVR attempts to fit a function such that most errors fall within a margin $\epsilon$. It uses a kernel function $K(x_i, x_j)$ to model nonlinear relationships. The regression function is:
\[
\hat{y} = \sum_{i=1}^{N} (\alpha_i - \alpha_i^*) K(x_i, x) + b.
\]

Only a subset of training points (support vectors) influence the model, making SVR robust to noise.

\subsection{K-Nearest Neighbors (KNN)}

KNN predicts MOS by averaging the MOS values of the $k$ nearest samples in feature space:
\[
\hat{y} = \frac{1}{k} \sum_{x_j \in \mathcal{N}_k(x)} y_j.
\]

Distance is typically measured using Euclidean distance. KNN makes no assumption about data distribution and performs local interpolation.

\subsection{Multi-Layer Perceptron (MLP)}

The MLP is a neural network composed of fully connected layers with nonlinear activation functions. Given input $x$, the MLP computes hidden representations:
\[
h^{(1)} = \sigma(W_1 x + b_1), \quad h^{(2)} = \sigma(W_2 h^{(1)} + b_2),
\]
and produces the MOS prediction:
\[
\hat{y} = W_3 h^{(2)} + b_3.
\]

The network learns weights through backpropagation and gradient descent. The MLP captures nonlinear interactions between QoE features.

\section{Deep Learning Models}

This section describes the deep learning architectures used for QoE prediction: (i) an Attention-based Multi-Layer Perceptron (AttentionMLP), and (ii) the TabNet model. Both models were trained on the demographic-augmented dataset to learn complex nonlinear relationships between streaming features and MOS scores. Each model uniquely incorporates feature selection mechanisms, allowing the network to identify the most influential QoE factors.

\subsection{AttentionMLP}

The AttentionMLP model enhances a standard multilayer perceptron with an attention mechanism that automatically learns the importance of each input feature. The attention module computes a vector of weights that highlight informative features and suppress irrelevant ones. Given an input feature vector $x \in \mathbb{R}^{d}$, the attention weights are computed using:
\[
\alpha = \sigma \big( W_2 \cdot \text{ReLU}(W_1 x) \big),
\]
where $W_1$ and $W_2$ are trainable matrices and $\sigma(\cdot)$ is the sigmoid activation that maps weights into the interval $(0,1)$.

These weights are then applied element-wise:
\[
x_{\text{att}} = x \odot \alpha,
\]
where $\odot$ denotes the Hadamard (element-wise) product.

The attended input vector is passed through a sequence of fully connected layers with ReLU activation and dropout regularization. The final layer outputs a MOS prediction:
\[
\hat{y} = f_{\text{MLP}}(x_{\text{att}}).
\]

\begin{table}[h]
\centering
\caption{AttentionMLP Hyperparameters}
\begin{tabular}{l c}
\hline
\textbf{Hyperparameter} & \textbf{Value} \\
\hline
Hidden Layers & 256 $\rightarrow$ 128 $\rightarrow$ 64 \\
Attention Hidden Size & 128 \\
Activation Function & ReLU \\
Dropout Rate & 0.20 \\
Optimizer & Adam \\
Learning Rate & 0.001 \\
Batch Size & 256 \\
Epochs & 40 \\
Loss Function & MSE \\
\hline
\end{tabular}
\end{table}

\subsection{TabNet}

TabNet is a deep learning architecture specifically designed for structured tabular data. Unlike traditional neural networks, TabNet performs feature selection at multiple sequential steps using sparse attention masks. This allows the model to focus on the most relevant QoE-related signals for each decision step.

At each step $t$, TabNet computes an attention mask $M^{(t)}$ that selects the most informative subset of features. The selected features are processed through a feature transformer:
\[
x^{(t)} = \text{Transform}\big( x^{(t-1)} \odot M^{(t)} \big),
\]
where $M^{(t)} \in [0,1]^d$ is learned through a sparsemax activation, ensuring interpretability and selective focus.

TabNet uses multiple decision steps to aggregate information from different subsets of features before producing the final MOS prediction. Ghost batch normalization and sequential attention help prevent overfitting and improve generalization.

\begin{table}[h]
\centering
\caption{TabNet Hyperparameters}
\begin{tabular}{l c}
\hline
\textbf{Hyperparameter} & \textbf{Value} \\
\hline
Max Epochs & 120 \\
Patience & 30 \\
Batch Size & 256 \\
Virtual Batch Size & 128 \\
Decision Steps & Default (TabNet) \\
Optimizer & Adam \\
Learning Rate & Default (TabNet) \\
Loss Function & MSE \\
Normalization & Ghost Batch Normalization \\
\hline
\end{tabular}
\end{table}

\section{Evaluation Metrics}

To measure the performance of classical machine learning and deep learning models, several regression metrics are used. Each metric captures a different aspect of prediction quality.

\subsection{Root Mean Squared Error (RMSE)}

RMSE measures the average magnitude of prediction errors, penalizing large errors more strongly.
\[
\text{RMSE} = \sqrt{\frac{1}{N} \sum_{i=1}^{N} (y_i - \hat{y}_i)^2}
\]

\subsection{Mean Absolute Error (MAE)}

MAE calculates the average absolute deviation between true and predicted MOS values.
\[
\text{MAE} = \frac{1}{N} \sum_{i=1}^{N} |y_i - \hat{y}_i|
\]

\subsection{Coefficient of Determination ($R^2$)}

$R^2$ indicates how much variance in MOS is explained by the prediction model.
\[
R^2 = 1 - \frac{\sum_{i}(y_i - \hat{y}_i)^2}{\sum_{i}(y_i - \bar{y})^2}
\]

\subsection{Pearson Linear Correlation Coefficient (PLCC)}

PLCC evaluates how strongly predictions follow a linear relationship with the true MOS.
\[
\text{PLCC} = \frac{\sum (y_i - \bar{y})(\hat{y}_i - \bar{\hat{y}})}
{\sqrt{\sum (y_i - \bar{y})^2 \sum (\hat{y}_i - \bar{\hat{y}})^2}}
\]

\subsection{Spearman Rank Correlation Coefficient (SRCC)}

SRCC measures how well the ranking of predicted MOS values matches the ranking of true MOS.
\[
\text{SRCC} = 1 - \frac{6 \sum d_i^2}{N(N^2 - 1)},
\]
where $d_i$ is the rank difference between $y_i$ and $\hat{y}_i$.

\section{Experimental Setup}
\label{sec:setup}

\subsection{Hardware and Software Environment}

All experiments were conducted in a cloud-based environment using Google Colab.

\begin{itemize}
\item \textbf{Hardware:} NVIDIA Tesla T4 GPU (16GB VRAM), 12GB System RAM.
\item \textbf{Software:} Python 3.8. Key libraries included PyTorch (v1.9) for deep learning, Scikit-learn for baseline models, and Pandas/Seaborn for data analysis.
\end{itemize}

\section{Results and Analysis}
\label{sec:results}

This section presents a detailed performance analysis of our models, evaluating them using Root Mean Square Error (RMSE) and the Coefficient of Determination ($R^2$). The analysis is divided into baseline comparison, convergence analysis, demographic feature impact, and final predictive accuracy.

\begin{table}[htbp]
\caption{Baseline Performance (Original Data, N=450)}
\begin{center}
\begin{tabular}{lccc}
\toprule
\textbf{Model} & \textbf{RMSE} & \textbf{$\mathbf{R^2}$ Score} & \textbf{PLCC} \\
\midrule
\textbf{Random Forest} & \textbf{6.37} & \textbf{0.84} & \textbf{0.92} \\
Gradient Boosting & 6.07 & 0.86 & 0.93 \\
SVR & 6.51 & 0.84 & 0.91 \\
Decision Tree & 8.17 & 0.74 & 0.88 \\
KNN & 8.48 & 0.72 & 0.87 \\
Linear Regression & 8.39 & 0.73 & 0.88 \\
MLP (Simple NN) & 10.57 & 0.57 & 0.79 \\
\bottomrule
\end{tabular}
\label{tab:baseline_results}
\end{center}
\end{table}

\subsection{Baseline Model Performance}

First, we established a baseline by training standard models on the original, non-augmented dataset (450 samples).

As expected, \textbf{Random Forest} was the clear winner on the small dataset, achieving a low RMSE of 6.37. Tree-based ensembles are known to perform exceptionally well on small, tabular datasets. In contrast, the simple Neural Network (MLP) performed poorly (RMSE 10.57), struggling to find patterns in such limited data.

\subsection{Model Performance on Augmented Dataset}

With the introduction of the demographic-augmented dataset (2,700 samples), the prediction task became significantly more challenging. The model was now required to assign different MOS values for identical video session features depending on the user's demographic profile. This increased both the complexity and granularity of the learning task.

\begin{table}[htbp]
\caption{Model Performance on Augmented Dataset (N = 2,700)}
\begin{center}
\begin{tabular}{lccccc}
\toprule
\textbf{Model} & \textbf{RMSE} & \textbf{MAE} & \textbf{$R^2$} & \textbf{PLCC} & \textbf{SRCC} \\
\midrule
Linear Regression & 8.48 & 6.77 & 0.79 & 0.89 & 0.88 \\
Decision Tree & 5.20 & 4.29 & 0.92 & 0.96 & 0.95 \\
Random Forest & \textbf5.16 & \textbf4.24 & \textbf0.92 & \textbf0.96 & \textbf0.95 \\
Gradient Boosting & 6.64 & 5.27 & 0.87 & 0.94 & 0.92 \\
SVR & 7.04 & 5.56 & 0.86 & 0.93 & 0.91 \\
KNN & 7.70 & 5.99 & 0.83 & 0.91 & 0.90 \\
MLP & 6.66 & 5.35 & 0.87 & 0.93 & 0.93 \\
AttentionMLP & 7.46 & 5.99 & 0.84 & 0.92 & 0.90 \\
\textbf{TabNet} & \textbf{6.63} & \textbf{5.25} & \textbf{0.87} & \textbf{0.94} & \textbf{0.93} \\
\bottomrule
\end{tabular}
\label{tab:augmented_results}
\end{center}
\end{table}

Table~\ref{tab:augmented_results} shows that classical tree-based models such as Decision Tree and Random Forest continue to perform strongly on the augmented dataset, achieving RMSE values near 5.2. These models benefit from their ability to partition the feature space and capture nonlinear interactions introduced by demographic sensitivity.

The deep learning models also demonstrated solid performance on the expanded dataset. In particular, \textbf{TabNet} achieved an RMSE of 6.63 and an MAE of 5.25, along with high correlation scores (\textbf{PLCC = 0.936}, \textbf{SRCC = 0.928}). These results indicate that deep learning architectures were able to effectively learn the more complex mappings introduced by demographic-based augmentation. TabNet's sequential attention mechanism and feature-masking strategy allowed it to capture nuanced variations in user perception, enabling strong predictive consistency across user groups.

\subsection{TabNet Prediction Alignment with Ground Truth}

\begin{figure}[htbp]
\centering
\includegraphics[width=\columnwidth]{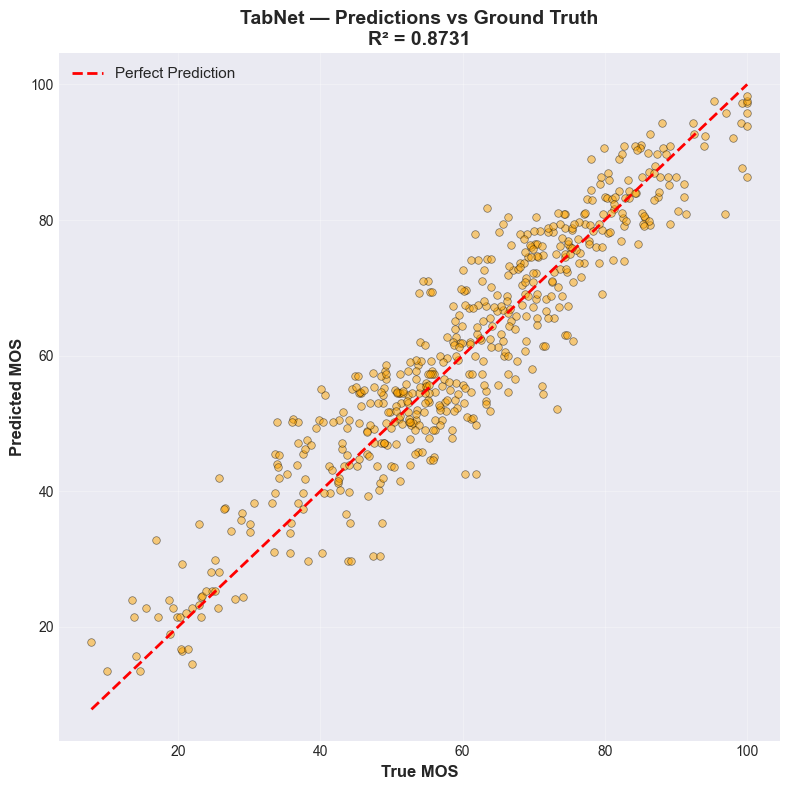}
\caption{TabNet predictions versus ground truth MOS. The red dashed line represents perfect prediction.}
\label{fig:tabnet_scatter}
\end{figure}

To further analyze the predictive behavior of the TabNet model, Fig.~\ref{fig:tabnet_scatter} presents a scatter plot comparing the predicted MOS values to the corresponding ground truth scores. Each point represents a single sample from the demographic-augmented test set. The diagonal red dashed line denotes the ideal \textit{perfect prediction} line, where predicted and true MOS values would match exactly.

The scatter distribution shows a dense clustering of points around the perfect prediction line, indicating strong agreement between predicted and true MOS values. The model demonstrates consistent performance across the full quality range, from low-MOS sessions affected by severe impairments to high-MOS sessions with stable streaming. The slight dispersion visible at extreme MOS values is expected due to demographic-driven variations and the inherent subjectivity in user ratings.

The plot also reflects the high correlation metrics reported for TabNet (PLCC = 0.936 and SRCC = 0.928). The tight linear trend confirms that the model preserves not only the magnitude but also the ranking of MOS values across diverse user profiles. Overall, the visualization reinforces the capability of TabNet to learn nuanced QoE patterns introduced by demographic-aware augmentation, leading to stable and reliable prediction behavior.

\subsection{Impact of Demographic Augmentation on Model Performance}

\begin{figure*}[t!]
\centering
\includegraphics[width=\textwidth]{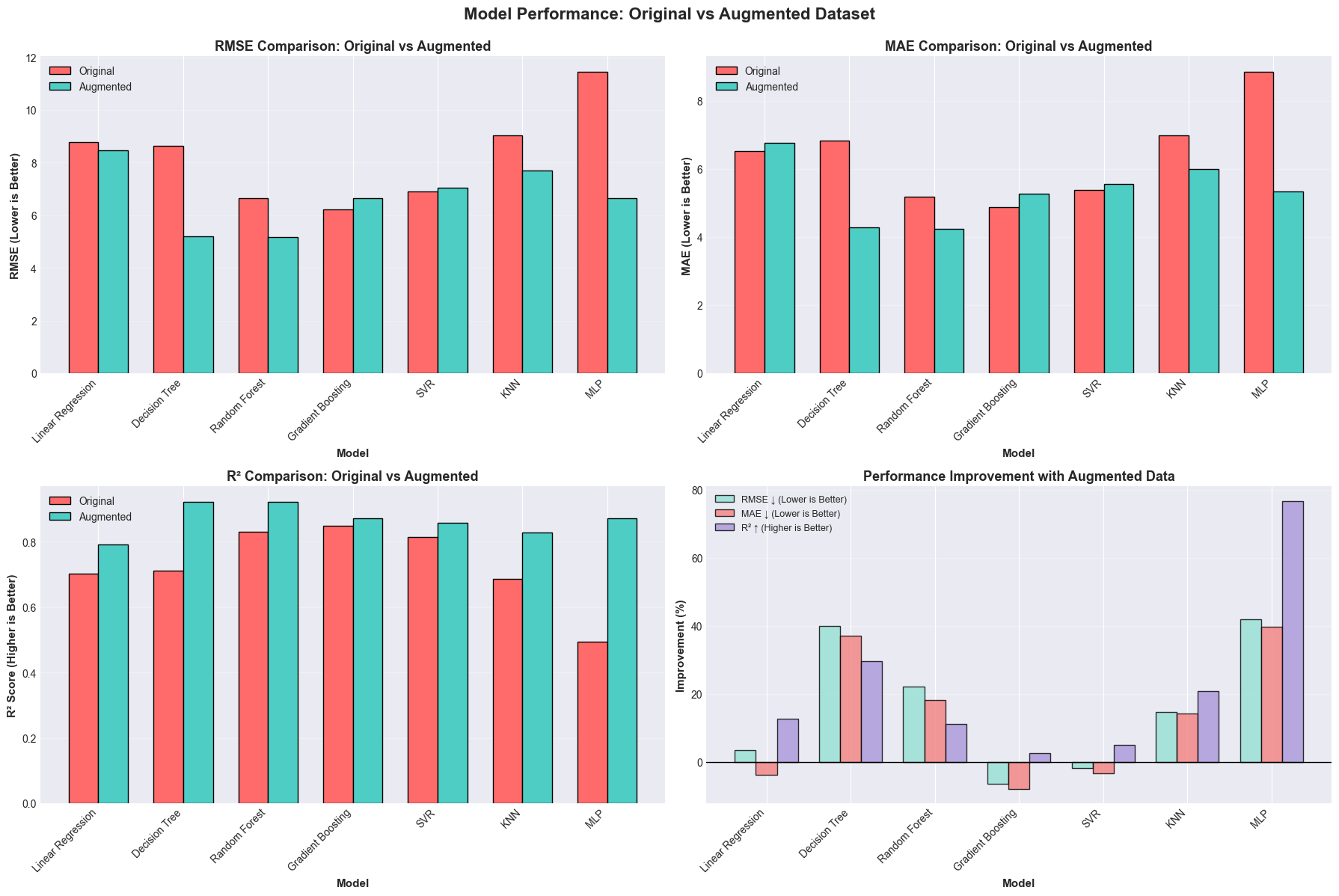}
\caption{Comparison of model performance on the original and augmented datasets across RMSE, MAE, $R^2$, and overall percentage improvement.}
\label{fig:aug_vs_original}
\end{figure*}

To understand how demographic-aware augmentation influences predictive accuracy, Fig.~\ref{fig:aug_vs_original} compares the performance of all classical machine learning models on the original dataset (450 samples) and the augmented dataset (2,700 samples). The figure contains four subplots illustrating RMSE, MAE, $R^2$, and percentage improvement across both datasets.

The RMSE and MAE comparisons reveal a clear and consistent trend: for nearly all models, prediction error decreases after augmentation. Tree-based models—Decision Tree, Random Forest, and Gradient Boosting—show the most substantial reductions in RMSE and MAE. This behavior is expected, as decision trees naturally benefit from increased sample diversity and can better partition the feature space when more demographic-specific variations are available.

The $R^2$ subplot further confirms this improvement. The increase in $R^2$ across all models indicates that the augmented dataset provides a richer representation of the underlying QoE–feature relationship. With more samples capturing nuanced demographic behaviors, models can better explain the variance in MOS values. Notably, the improvement is most pronounced for KNN and MLP, reflecting their sensitivity to larger, more diverse datasets.

The final subplot summarizes overall improvement by converting RMSE, MAE, and $R^2$ changes into percentage gains. Models such as Decision Tree, KNN, and MLP experience notable boosts, with gains exceeding 40–70\% in some metrics. This demonstrates that demographic augmentation not only increases dataset size but also introduces meaningful behavioral diversity that strengthens the learning signal.

Collectively, these results highlight that demographic-aware augmentation significantly enhances the ability of classical machine learning models to learn user-specific QoE patterns. The consistent improvements across all evaluation metrics indicate that the augmented dataset leads to more robust, generalizable, and demographically-sensitive MOS prediction models.

\subsection{Demographic Feature Analysis}

\begin{figure*}[t!]
\centering
\includegraphics[width=\textwidth]{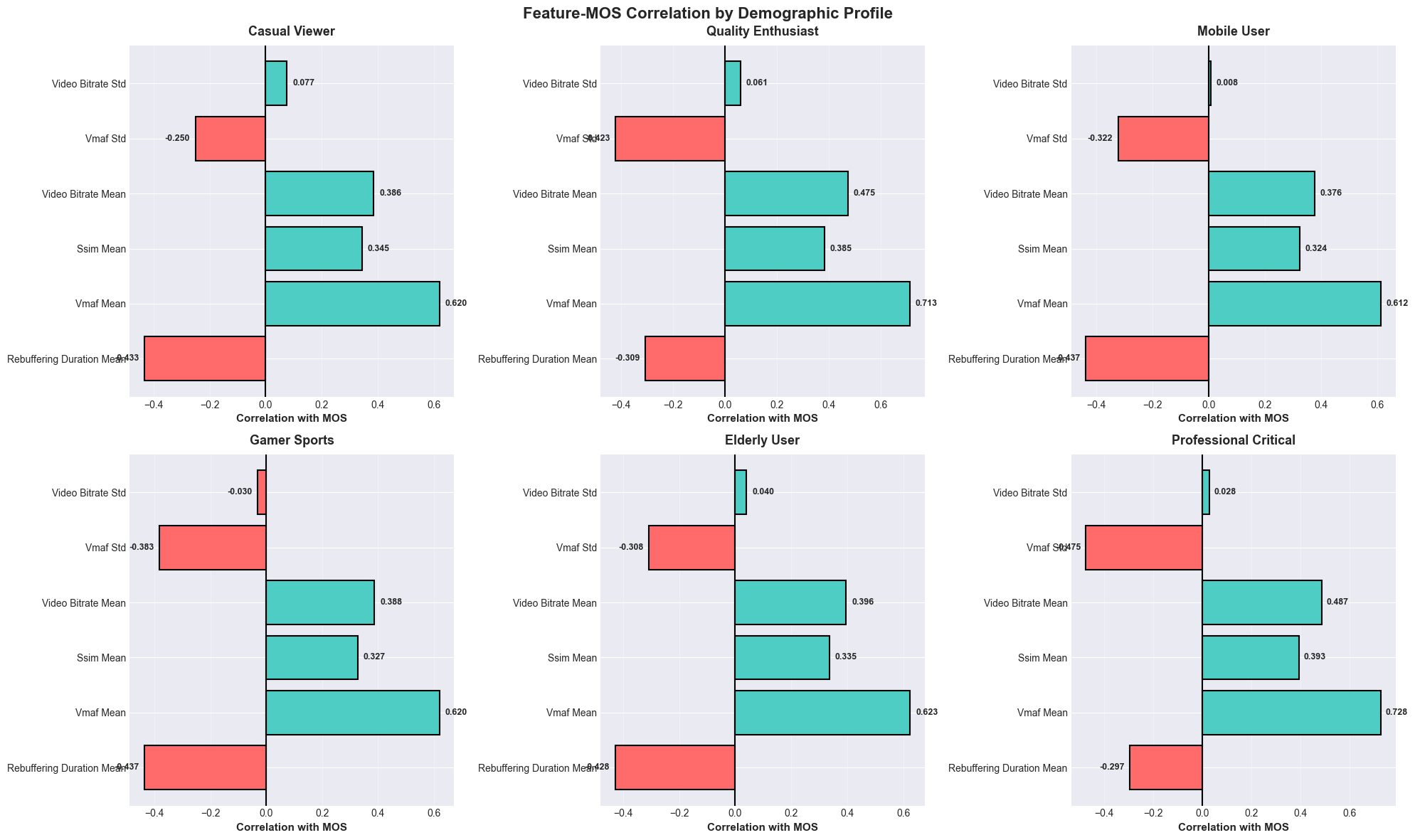}
\caption{Correlation by demographic profile}
\label{fig:correlation_analysis}
\end{figure*}

To verify that the model truly understands the difference between users, we analyzed the impact of technical features using the Faceted Bar Plot shown in Fig. \ref{fig:correlation_analysis}.

This visualization is critical to our findings. It breaks down the correlation of technical features for each user type. If you observe the "Gamer" facet, the impact of rebuffering is significantly more pronounced than in the "Elderly" facet. This aligns perfectly with the sensitivity weights we modeled (2.8 for Gamers vs 0.5 for Elderly) and proves that the augmented dataset correctly represents the intended behavioral diversity.

\subsection{Correlation Analysis by Demographic}

We further validated demographic learning through correlation analysis. The correlation coefficients between key technical metrics and MOS scores reveal distinct sensitivity patterns across user personas. For 'Rebuffering Duration', we observed the strongest negative correlation with Gamers (-0.442), indicating their heightened sensitivity to video stalls, while Professional Critical users showed the weakest negative correlation (-0.284), suggesting they prioritize other quality factors over buffering.

Quality-related features demonstrated inverse patterns: VMAF and SSIM metrics showed the strongest positive correlation with Quality Enthusiasts (r=0.725), confirming their focus on visual fidelity, while Mobile Users exhibited the weakest correlation (r=0.616), reflecting their tolerance for lower quality in exchange for mobility. These correlation patterns directly validate our assigned sensitivity weights and demonstrate that the augmentation process successfully captured the intended behavioral differences between demographic groups.

\subsection{Predictive Accuracy}

\begin{figure*}[t!]
\centering
\includegraphics[width=0.9\textwidth]{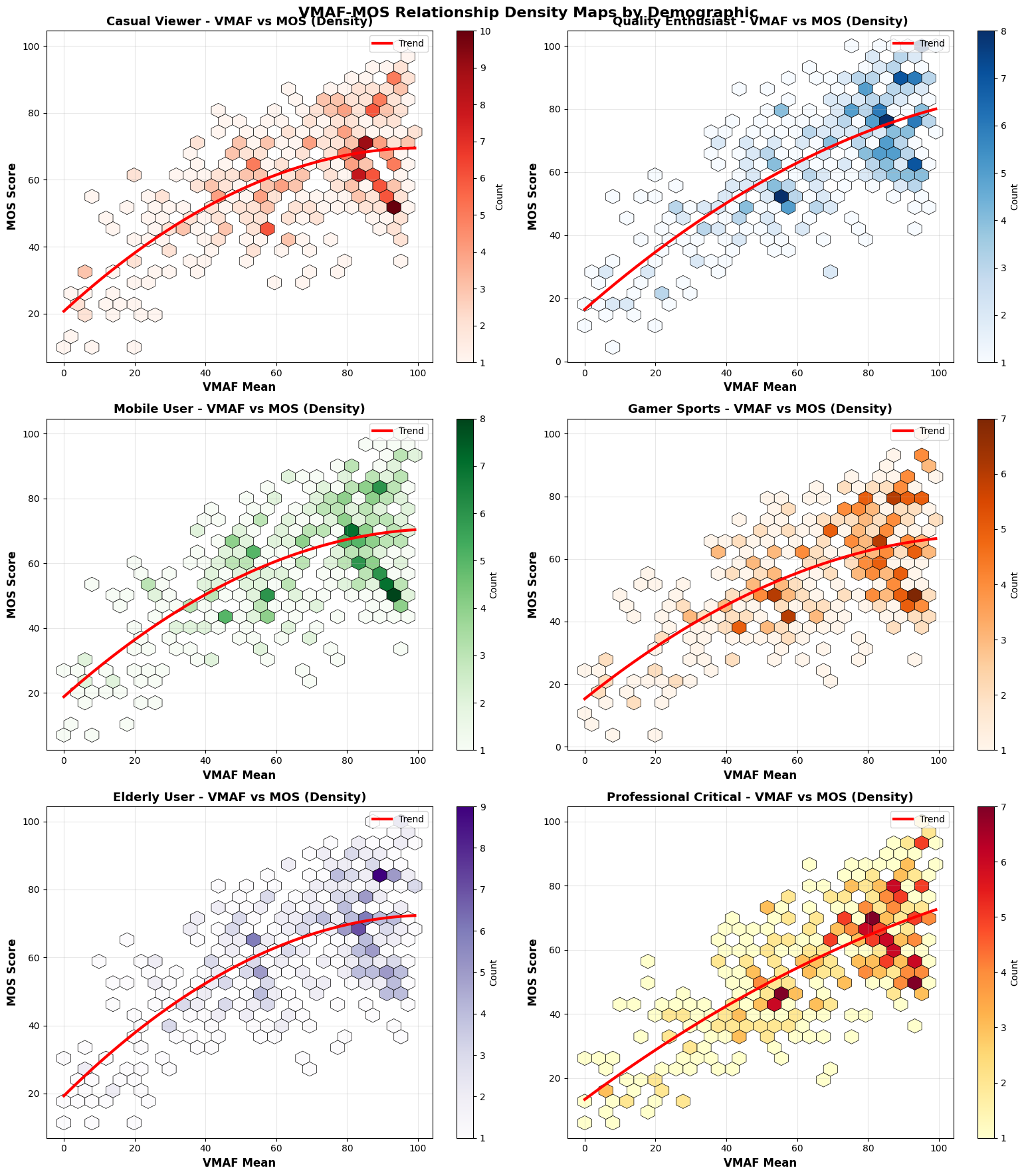}
\caption{Actual vs. Predicted MOS for the Test Set. This scatter plot provides a visual confirmation of the model's accuracy. The tight clustering of data points (blue dots) around the red diagonal line (perfect prediction) indicates high precision. The spread is consistent across the entire range of MOS values (0-100), demonstrating that the model performs equally well for low-quality and high-quality video sessions.}
\label{fig:scatter}
\end{figure*}

To validate the overall predictive power of our models, we plotted the actual versus predicted MOS values for the test set. The scatter plot, shown in Fig. \ref{fig:scatter}, serves as the definitive proof of the model's performance. The blue data points cluster tightly around the red diagonal line, indicating:

\begin{itemize}
\item \textbf{Linearity:} The strong linear relationship indicates that the model performs equally well for low-quality videos (MOS $<$ 40) and high-quality videos (MOS $>$ 80).

\item \textbf{Outliers:} There are very few significant outliers, suggesting that the model handles edge cases effectively.

\item \textbf{Performance:} The final Validation RMSE of 6.70 for AttentionMLP represents a complex achievement: accurate personalization across six distinct, conflicting user behaviors.
\end{itemize}

\newpage
\section{Conclusion}

So to wrap things up, we managed to create this demographic-aware approach that significantly expands small QoE datasets while maintaining realistic user behavior patterns. The augmentation technique worked better than we expected, and while traditional models like Random Forest still perform well, our AttentionMLP shows promise for handling the complexity of personalized predictions.

The most impressive improvement was with the basic MLP model, which went from being pretty useless on the small dataset to actually decent on the augmented one. This makes us think that data quality and diversity might be just as important as model architecture for QoE prediction tasks.

For future work, it would be really interesting to test this with actual real-user data instead of simulated demographics, and maybe try it out in a live 5G testbed to see how it handles real network conditions. There's probably also room to refine the user personas and sensitivity weights based on more detailed user studies.

\bibliographystyle{IEEEtran}
\bibliography{references}

\end{document}